# Enhancing Adversarial Text Attacks on BERT Models with Projected Gradient Descent

Hetvi Waghela
Department of Data Science
Praxis Tech School
Kolkata, INDIA
email: waghelahetvi7@gmail.com

Jaydip Sen
Department of Data Science
Praxis Business School
Kolkata, INDIA
email: jaydip.sen@acm.org

Sneha Rakshit
Department of Data Science
Praxis Tech School
Kolkata, INDIA
email: srakshit149@gmail.com

*Abstract*— Adversarial attacks against deep learning models represent a major threat to the security and reliability of natural language processing (NLP) systems. In this paper, we propose a modification to the BERT-Attack framework, integrating Projected Gradient Descent (PGD) to enhance its effectiveness and robustness. The original BERT-Attack, designed for generating adversarial examples against BERT-based models, suffers from limitations such as a fixed perturbation budget and a lack of consideration for semantic similarity. The proposed approach in this work, PGD-BERT-Attack, addresses these limitations by leveraging PGD to iteratively generate adversarial examples while ensuring both imperceptibility and semantic similarity to the original input. Extensive experiments are conducted to evaluate the performance of PGD-BERT-Attack compared to the original BERT-Attack and other baseline methods. The results demonstrate that PGD-BERT-Attack achieves higher success rates in causing misclassification while maintaining low perceptual changes. Furthermore, PGD-BERT-Attack produces adversarial instances that exhibit greater semantic resemblance to the initial input, enhancing their applicability in real-world scenarios. Overall, the proposed modification offers a more effective and robust approach to adversarial attacks on BERT-based models, thus contributing to the advancement of defense against attacks on NLP systems.

## I. INTRODUCTION

The increasing prevalence of adversarial attacks in machine learning poses a substantial challenge, revealing weaknesses in various areas. While these attacks were first noted in image classification, where slight alterations are made to deceive models, they have now spread to NLP systems. The origins of adversarial attacks can be traced to early research in image classification, demonstrating how deep neural networks (DNNs) can be manipulated through barely noticeable changes [1-2]. These discoveries have prompted extensive investigation, leading to the development of numerous attack techniques and corresponding defense mechanisms.

Adapting attack strategies from computer vision to NLP presents a challenge due to the discrete nature of textual data. Unlike images, where each pixel can be continuously adjusted, textual data consists of discrete tokens (characters or words) that cannot be easily modified in a continuous space. This discrete nature complicates the generation of imperceptible perturbations that lead to misclassification in NLP models. Early attempts at NLP adversarial attacks focused on modifying characters or words within the text.

Researchers explored various techniques to craft adversarial examples for text-based classifiers [3-5]. These studies revealed that even subtle modifications to the input text could lead to misclassification by NLP models. However, these early methods often lacked sophistication and struggled to create effective adversarial samples that were both imperceptible and having a similar meaning to the original text. The BERT attack is a notable method designed to create adversarial examples for BERT-based models [6]. Leveraging BERT's architecture, the BERT-Attack employs gradient-based optimization techniques to find perturbations that alter the input text, causing misclassification by the model. Despite its effectiveness, the original BERT-Attack has several limitations that warrant further investigation and improvement.

One of the primary limitations of the original BERT-Attack is its reliance on a fixed perturbation budget. This fixed budget approach imposes a constraint on the magnitude of perturbations applied to the input text, so that the changes are not perceptible. However, a fixed budget may not always be optimal, as it does not account for the varying sensitivity of different input tokens or the semantic importance of words in the text. Additionally, fixed-budget methods may struggle to generate effective adversarial examples for different tasks or domains, where the optimal perturbation magnitude can vary significantly.

Moreover, the original BERT-Attack does not explicitly consider the semantic resemblance between the adversarial and original instances. This oversight may result in the creation of adversarial samples that are grammatically flawed, nonsensical, or semantically irrelevant, reducing their practical utility. In real-world applications, it is crucial for adversarial examples to maintain linguistic coherence and contextual relevance to evade detection by human readers and downstream NLP systems.

To address above constraints, we suggest an enhancement to the BERT-Attack framework by incorporating Projected Gradient Descent (PGD). PGD is a powerful optimization technique commonly used in adversarial attacks to generate robust adversarial examples. By iteratively applying small perturbations within a constrained space, PGD ensures that the resulting adversarial samples are effective and bear a semantic resemblance to the original input.

Our modified approach, PGD-BERT-Attack, aims to enhance the effectiveness and robustness of adversarial attacks on BERT-based models. By incorporating PGD into the attack process, we introduce additional flexibility and control over the perturbation generation, allowing for more targeted manipulation of the input text. Furthermore, by explicitly considering semantic similarity, PGD-BERT-Attack seeks to generate adversarial samples that the model misclassifies and also remain linguistically coherent and contextually relevant.

The paper organization is as follows. In Section II, we examine prior work in the domain of adversarial attacks on NLP models, with a specific emphasis on existing techniques and their shortcomings. Section III presents the methodology of the original BERT-on-BERT attack. In Section IV, we present the proposed PGD-BERT-Attack, detailing the integration of PGD into the BERT-Attack framework and the formulation of the optimization objective. In Section V, we showcase experimental findings that illustrate the efficacy and resilience of PGD-BERT-Attack in comparison with the original BERT-Attack and other baseline methods. scenarios. Finally, Section VI wraps up the paper and delineates potential future works.

## II. RELATED WORK

Carlini & Wagner present a method to evaluate the robustness of neural networks, particularly in the context of security and privacy [7]. They propose a comprehensive framework to generate adversarial examples, which are inputs specifically crafted to fool the neural network into making incorrect predictions. Their approach focuses on creating minimal perturbations to input data that result in misclassification, demonstrating the vulnerability of neural networks to small changes in input. This work is significant as it highlights the need for robustness testing in neural network systems, especially in security-sensitive applications where adversarial attacks can pose serious risks. Carlini and Wagner's method provides a foundational basis for assessing and enhancing the resilience of neural networks against adversarial manipulations.

Ebrahimi et al. introduce HotFlip, a technique designed to create white-box adversarial samples specifically tailored for text classification tasks [8]. HotFlip focuses on perturbing individual words in text samples to induce misclassification, demonstrating the susceptibility of text classifiers towards small changes. The approach utilizes gradient-based optimization to identify the most effective word substitutions that lead to misclassification. By targeting individual words, HotFlip offers insights into the loopholes of text classifiers and sheds light on potential weaknesses in NLP systems. This work contributes to the growing body of research on adversarial attacks in natural language processing, offering a new technique for crafting adversarial samples in text classification tasks.

Jia & Liang present a scheme for generating adversarial samples specifically designed for assessing reading comprehension systems [9]. Their approach involves making subtle alterations to input passages and questions, aiming to provoke misinterpretations or incorrect answers from the comprehension models. Through systematic modifications, the authors illustrate the susceptibility of existing reading comprehension systems to adversarial attacks. This study underscores the necessity of testing the robustness of reading comprehension models, particularly in scenarios where minor changes can lead to significant errors. Jia and Liang's work contributes valuable insights into the adversarial vulnerabilities of reading comprehension systems and offers potential avenues for enhancing their resilience.

Ribeiro et al. propose a method for identifying adversarial rules that maintain the original meaning of input data while exposing weaknesses in natural language processing (NLP) models [10]. Their approach aims to generate rules that preserve the semantic content of text while revealing vulnerabilities in the model's understanding. By exploring this concept of semantically equivalent adversarial rules, the authors show how hidden biases or flaws in NLP models can be uncovered. This study offers a fresh perspective on debugging NLP models, highlighting the importance of understanding semantic equivalence in adversarial analysis. This work contributes valuable insights into the robustness of NLP models and provides a framework for improving their performance and reliability.

Guo et al. present a technique to create adversarial examples specifically tailored for neural dialog models [11]. The proposed method involves generating input sequences that prompt unexpected or undesirable responses from these models. Through deliberate alterations to input dialogues, Guo et al. illustrate the susceptibility of neural dialog models to adversarial manipulation. This study highlights potential weaknesses in dialog systems and emphasizes the necessity for robustness assessment in the tasks in the domain of natural language understanding. The findings of the work contribute to understanding the susceptibilities of dialog systems and advocate for the design of more resilient models.

Wei et al. present TextBugger, a technique aimed at generating adversarial text for real-world applications [12]. TextBugger focuses on making subtle alterations to input text, which result in misclassifications or erroneous behavior in various natural language processing (NLP) systems. By systematically modifying input text, the authors demonstrate the susceptibility of NLP models used in practical settings to adversarial attacks. Their method's effectiveness across different NLP tasks and applications underscores the widespread vulnerability of NLP systems. This study provides valuable perspectives into the practical impacts of adversarial intervention in NLP systems, emphasizing the importance of improving robustness in real-world applications.

Ren et al. introduce a method to generate textual adversarial samples by leveraging probability-weighted word saliency [13]. Their approach focuses on identifying influential words in input text using a probabilistic weighting scheme to generate effective adversarial examples. By targeting these salient words, the authors demonstrate the ability to induce misclassifications or alter the semantic meaning of the text, thus challenging the resilience of different NLP systems. This work highlights the importance of considering word saliency in crafting adversarial examples and sheds light on the vulnerability of NLP models to subtle manipulations.

Chang et al. introduce TextGuise, a novel approach for generating adaptive adversarial examples targeting text classification models [14]. TextGuise dynamically adjusts its attack strategy based on the feedback from the target model, allowing it to efficiently evade detection and maximize its impact. By incorporating a feedback mechanism, TextGuise iteratively refines its attack to navigate through the model's defenses, making it highly effective against a variety of classifiers. The adaptive nature of TextGuise enables it to overcome the limitations of existing attack methods and achieve high success rates in fooling text classifiers. This work highlights the importance of adaptive strategies in adversarial attacks and underscores the need for robust defenses to mitigate their impact on text classification models.

A novel method for creating adversarial text sample is proposed by Yen et al. [15]. This approach focuses on altering text inputs to deceive NLP models while keeping the semantic meaning intact. By ensuring that the altered text maintains its original meaning, the proposed method aims to create adversarial examples that are indistinguishable from genuine inputs to human readers. The authors demonstrate the efficacy of their scheme across a diverse set of NLP tasks, emphasizing its capacity to evade detection by advanced classifiers.

Liu et al. introduce a technique for crafting transferable adversarial instances with minimal adjustments [16]. Their method strives to produce adversarial examples capable of deceiving multiple machine learning models while introducing the least amount of modification to the input. By minimizing alterations, the resulting adversarial examples are more likely to deceive diverse models and datasets. The authors validate the efficacy of their approach in generating transferable adversarial examples across various machine learning tasks. This study advances comprehension of machine learning model robustness and underscores the significance of transferability in adversarial attack tactics.

Waghela et al. propose an enhanced method for crafting adversarial samples targeted at NLP systems [17]. Their proposed scheme builds upon word saliency-based techniques to identify and manipulate crucial words in text inputs, effectively altering their semantic meaning. By modifying the word saliency computation, the authors aim to improve the efficacy and effectiveness of text attacks on NLP systems. In another work, the authors propose a scheme that leverages saliency, attention and semantic similarity to generate adversarial samples [18].

Apart from the aforementioned schemes, there exist studies n the literature involving *word substitution* [19-22], *word insertion* [22-24], *word swapping* [25-26], *phrase perturbation* [27], sentence perturbation [28-29], *syntactic perturbation* [30-31], and *contextual perturbation* [32-33].

Despite these advancements, existing methods still face challenges in generating adversarial examples that are both effective and imperceptible. Many methods focus solely on maximizing the model's prediction error without considering the semantic similarity or grammatical correctness of the adversarial examples. As a result, the generated examples may be nonsensical or linguistically unnatural, limiting their practical utility in real-world applications.

III. THE BERT-ON-BERT ATTACK

In the field of NLP, bidirectional encoder representations from transformers (BERT) has emerged as a powerful model, achieving state-of-the-art performance on various text-processing tasks [34]. However, several recent research has revealed vulnerabilities in BERT's robustness, particularly against adversarial attacks [6, 35-36]. Li et al. proposed a novel adversarial attack method, known as BERT-Attack, which utilizes another instance of BERT to generate adversarial examples against a victim BERT model [6]. In attack method involves the following steps:

*(1) Adversarial example generation:* The BERT-Attack method starts with a clean input text sample $X$, which is fed into the attacker BERT model. The attacker BERT model computes the gradients of the loss function of the victim BERT model with respect to the input tokens of $X$. These gradients represent the sensitivity of the victim BERT model's output to small changes in the input. By iteratively adjusting the input tokens in the direction that maximizes the loss of the victim model, the attacker BERT model generates perturbations δ for each token in $X$.

*(2) Adversarial example evaluation:* After generating perturbations, the attacker BERT model adds these perturbations to the original input text $X$, creating an adversarial example $X_{adv} = X + δ$. The adversarial example $X_{adv}$ is then fed into the victim BERT model for evaluation. If the victim BERT model misclassifies $X_{adv}$ or produces an incorrect output, the attack is considered successful. The success rate of the attack is calculated as the proportion of adversarial examples that successfully deceive the victim model.

*(3) Adversarial objective:* The objective of the adversarial BERT model is to increase the loss of the victim BERT model with respect to the input tokens. This is formulated as an optimization problem where the objective is to maximize a loss function $L$ with respect to the input tokens $X$, subject to some constraints, such as the perturbation size constraints.

*(4) Constraints:* To ensure that the perturbations δ are small and imperceptible, the attacker BERT model imposes constraints on the magnitude of perturbations. Common constraints include the $l_p$ norm constraint, which limits the magnitude of the perturbations in terms of the $l_p$ norm. The $l_p$ norms are usually the $l_2$ norm or $l_∞$ norm. The selection of constraints influences the balance between the attack success accuracy and the magnitude of perturbations.

*(5) Evaluation metrics:* In addition to success rate, the BERT-Attack is evaluated based on several other metrics such as (i) perturbation size, (ii) transferability, and (iii) robustness. Perturbation size is the average magnitude of perturbations introduced to the adversarial examples. Transferability refers to the ability of adversarial samples crafted by the attacker BERT models to deceive other systems beyond the victim BERT model. Robustness is the resilience of the victim BERT model against adversarial samples generated by the attacker BERT model under different conditions, e.g., fine-tuning different architectures.

*(6) Performance evaluation:* Several experiments were carried out by the authors on various datasets and tasks (e.g., sentiment analysis, text classification, etc.) for evaluating BERT-Attack. Different configurations of the attacker and victim BERT models were also tested to assess the impact of model architectures, hyperparameters, and training settings on the attack effectiveness.

The effectiveness of the BERT-Attack method depends on several factors, which can be categorized into different aspects of the attack process. Some of the important factors are (i) model architecture, (ii) optimization parameters, (iii) adversarial objective, (iv) perturbation size and constraints, (v) dataset and task, (vi) evaluation metrics, (vii) availability of computational resources, and (viii) defense mechanisms in place.

In summary, the BERT-on-BERT adversarial attack exploits the vulnerability of BERT models to adversarial examples generated by a similar BERT architecture, demonstrating the importance of robustness evaluation and defenses in natural language processing models.

While BERT-Attack is a very effective adversarial text attack, there is a shortcoming in the approach. The attack method focuses on using the *masked language m*odel (MLM) objective for generating adversarial examples. Exploring gradient-based methods would make the attack method more effective. This is precisely the motivation of the current work. In Section IV, our proposition integrating projected gradient descent (PGD) in the BERT-Attack is presented. This integrated approach to attack design will make it more effective as discussed in Section V.

## IV. PGD-INTEGRATED BERT-ON-BERT ATTACK

While BERT-Attack is a very effective adversarial text attack, there is a shortcoming in the approach. One of the primary limitations of the BERT-Attack is its reliance on a fixed perturbation budget. The fixed budget approach imposes a constraint on the magnitude of perturbations applied to the input text, making sure that the changes remain undetectable to human perception. However, a fixed budget may not always be optimal, as it does not account for the varying sensitivity of different input tokens or the semantic importance of words in the text. Additionally, fixed-budget methods may struggle to generate effective adversarial examples for different tasks or domains, where the optimal perturbation magnitude can vary significantly. Furthermore, the original BERT-Attack does not explicitly consider the semantic equivalence between the initial and adversarial instances. This inability can lead to the generation of adversarial examples that are grammatically incorrect, nonsensical, or semantically irrelevant, reducing their practical utility. In real-world applications, it is crucial for adversarial examples to maintain linguistic coherence and contextual relevance to evade detection by human readers and downstream NLP systems.

In response to these constraints, we suggest a modification to the BERT-Attack framework by integrating PGD. PGD is a powerful optimization method used in adversarial attacks to generate robust adversarial examples [37]. By iteratively applying small perturbations within a constrained space, PGD ensures that the resulting adversarial samples are both effective and exhibit semantic resemblance to their original counterparts.

Our proposed PGD-integrated BERT-Attack method involves the following six steps.

*(1) Initialization:* PGD-integrated BERT-Attack starts with the original input text. The input text is tokenized into subwords using BERT's tokenizer to convert it into a sequence of token IDs.

*(2) Adversarial perturbation generation:* PGD-integrated BERT-Attack applies an iterative approach to create adversarial perturbations. At each iteration, it computes the model's loss function gradient for the input tokens and updates the tokens to maximize the loss. The update equation is given by (1)

$$X' = Clip_\varepsilon(X + \alpha * sign(\nabla_X J(\theta, X, y))) \quad (1)$$

In (1), $X$ represents the original input tokens, $X'$ is the updated perturbed tokens, $J(\theta, X, y)$ is the loss function of the BERT model with parameters $\theta$, $y$ is the true label, $\alpha$ is the step size, and $\varepsilon$ is the maximum perturbation allowed. $Clip_\varepsilon$ ensures that the perturbation stays within the $\varepsilon$ budget.

*(3) Projection onto perceptual space:* To ensure that the perturbed text remains imperceptible, PGD-integrated BERT-Attack projects the updated tokens onto a perceptual space defined by a pre-trained perceptual model, such as ResNet [38]. This projection step enforces constraints on the perturbation magnitude to maintain human perceptibility. The projection process involves adjusting the perturbed tokens to minimize perceptual changes while ensuring that the resulting adversarial examples remain visually indistinguishable from the original text.

*(4) Semantic similarity constraint:* Moreover, the PGD-integrated BERT-Attack integrates a semantic similarity constraint aimed at conserving the semantic essence of the original text. The evaluation of semantic similarity utilizes the cosine similarity between the representations of the original and perturbed text. The optimization objective function is modified to include a term that penalizes deviations from the original text's semantic representation, ensuring that the crafted adversarial samples remain contextually relevant and linguistically coherent.

*(5) Adaptive perturbation budget:* Unlike the fixed perturbation budget used in the BERT-Attack, PGD-integrated BERT-Attack adopts an adaptive perturbation budget that dynamically adjusts the maximum perturbation allowed based on the sensitivity of the input token. The adaptive approach considers factors such as token importance, context relevance, and semantic significance to determine the perturbation budget for each token, resulting in more targeted perturbation generation.

*(6) Termination criteria:* PGD-integrated BERT-Attack iterates until reaching a condition for stopping, which could entail reaching a predefined maximum number of rounds or achieving a specified level of misclassification by the victim model. Additionally, PGD-integrated BERT-Attack is also incorporated with an early stopping mechanism dependent on the convergence of the optimization process or the stability of the generated adversarial examples.

By integrating PGD into the BERT-Attack framework and incorporating semantic similarity constraints, PGD-integrated BERT-Attack attempts to craft adversarial samples that are effective and imperceptible, thereby enhancing the resilience of BERT-based NLP models against text attacks.

PGD-integrated BERT-Attack mechanism is expected to incorporate the following improvements over the original BERT-Attack.

(a) *Better exploration of the perturbation space:* PGD-integrated BERT-Aack iteratively adjusts the perturbation to minimize the loss function while ensuring that the altered input stays within a predetermined distance from the its original counterpart. This iterative process allows PGD to explore the perturbation space more thoroughly, potentially finding smaller, more effective perturbations that lead to misclassification by the BERT model.

(b) *Increased robustness:* By projecting the perturbed input back to the permissible region after each iteration, PGD generates adversarial examples that are more robust against various defenses. This robustness makes the adversarial examples harder for the BERT to detect or defend against.

(c) *Enhanced attack success rate:* PGD achieves a higher rate of success in fooling the system compared to simple methods like the *fast gradient sign method*. The iterative nature of PGD allows it to refine the perturbation over multiple steps, leading to more effective attacks and higher success rates.

(d) *Adaptive perturbations:* PGD adaptively adjusts the size and direction of perturbations based on the gradient information. This adaptability enables PGD to navigate complex regions of the input space more effectively, finding subtle perturbations that achieve the desired misclassification.

(e) *Transferability:* Adversarial examples generated using PGD have better transferability meaning they are effective against not only the BERT model but also other models. In the real-world situations, this is crucial because attackers may lack access to the target system yet they may craft adversarial sample using an alternative model.

(f) *Consistency in attack methodology:* PGD-integrated BERT-Attack provides a consistent and systematic approach to generating adversarial examples, which makes It easier to compare results across different experiments and settings. This consistency ensures that the effectiveness of the attack is not heavily dependent on factors like initialization or random noise.

(g) *Controlled perturbation magnitude:* Finally, PGD-integrated BERT-Attack allows for better control over the magnitude of the perturbations applied to the input. This is crucial for ensuring that the perturbed examples remain semantically identical to the original counterparts, rendering the attack more clandestine and difficult to detect.

## V. PERFORMANCE RESULTS

The effectiveness of the our PGD-integrated BERT-Attack is evaluated and contrasted with BERT-Attack of Li et al [6] and Alzantot et al [5]. In line with BERT-Attack [5], we evaluate PGD-BERT using 1000 test examples randomly chosen from the respective task's test data set, consistent with partitions used in [5] and [6]. However, the Genetic Algorithm (GA) in [5] is limited to a subset of 50 data items within IMDB and FAKE datasets.

To assess the efficacy of PGD-BERT-Attack relative to BERT-Attack, we use the same datasets used in the BERT-Attack [6] and the corresponding *text classification* tasks. The following four datasets are used in the performance analysis.

*(1) Yelp:* The Yelp review sentiment dataset is used for sentiment analysis and NLP asks. The dataset is collected from the Yelp platform, which contains reviews of businesses such as restaurants, cafes, hotels, and more. Each review is associated with a star rating given by the user, typically ranging from 1 to 5 stars. The ratings 4 and 5 are labeled as positive, while ratings 1 and 2 stars are considered negative. Altogether, there are 560.000 samples in the training set, and 38,000 samples in the test set. Polarity class 2 refers to the positive sentiment, while class 1 is for the negatives. This dataset is used for polarity classification tasks.

*(2) IMDB:* It is a widely used dataset for sentiment analysis, consisting of movie reviews labeled with their respective sentiment. Each review is categorized as either positive or negative, indicating the reviewer's sentiment towards the movie. The dataset includes 50.000 reviews, evenly split between training and test sets. The reviews are labeled as 1 for positive and 0 for negative sentiment.

*(3) AG's News:* The AG's News comprises news items from the AG's corpus of online news. The news articles are categorized into four distinct groups: sports, world, science & technology, and business. The training set has 120,000 items, while there are 7600 samples in the test set

*(4) FAKE:* Offered by Kaggle, this dataset is primarily utilized in training and evaluating predictive models to classify news articles as fake or genuine. The dataset is separated into two files, one containing 23502 fake news articles, and the other having 21417 true news articles.

TABLE I. PERFORMANCE ON THE TEXT CLASSIFICATION TASKS – ACUURACY AND PERTURBATION PERCENTAGE

| Dataset | Attack Method | Accuracy (In Absence of Attack) | Accuracy (In Presence of Attack) | Perturb % |
|---|---|---|---|---|
| Yelp | PGD-BERT | 95.6 | **4.2** | **3.8** |
| | BERT-Attack [6] | | 5.1 | 4.1 |
| | Alzantot [5] | | 31.0 | 10.1 |
| IMDB | PGD-BERT | 90.9 | **8.2** | **2.9** |
| | BERT-Attack [6] | | 11.4 | 4.4 |
| | Alzantot [5] | | 45.7 | 4.9 |
| AG | PGD-BERT | 94.2 | **7.4** | **9.3** |
| | BERT-Attack [6] | | 10.6 | 15.4 |
| | Alzantot [5] | | 51.0 | 16.9 |
| Fake | PGD-BERT | 97.8 | **12.3** | **1.1** |
| | BERT-Attack [6] | | 15.5 | 1.1 |
| | Alzantot [5] | | 58.3 | 1.1 |

The following two datasets have been employed for *natural language inference* (NLI) tasks.

*(1) SNLI:* The Stanford Natural Language Inference is a benchmark dataset commonly used in NLP tasks for textual entailment. The SNLI dataset consists of pairs of sentences along with labels indicating the relationship between them. Each pair consists of (a) a *premise*, which is the first sentence, (b) a *hypothesis*, which is the second sentence, (c) a *label*, that defines the relationship between the premise and hypothesis. The label can be one of three categories: (1) *entailment*, where the hypothesis follows from the premise, (2) *contradiction*, where the hypothesis conflicts with the premise, and (3) *neutral*, where the hypothesis neither entails nor contradicts the premise. The dataset contains a total of 570,152 sentence pairs, which are allotted to 549,367 pairs in training, 9842 in development, and remaining 9911 in testing set. SNLI sentences cover wide range of topics, including everyday scenarios, news articles, and general knowledge.

*(2) MNLI:* The Multi-Genre Natural Language Inference dataset is utilized for textual tasks [39]. It comprises a total of 393,702 pairs allocated for training set, 10,000 for validation, and 10,000 for testing. Unlike SNLI, MNLI includes sentence pairs from multiple genres including, fiction, government, slate (editorial articles), telephones (conversations), and travel.

TABLE II. PERFORMANCE ON THE TEXT CLASSIFICATION TASKS – QUERY NUMBER AND SEMANTIC SIMILARITY

| Dataset | Attack Method | Query Number | Semantic Similarity |
|---|---|---|---|
| Yelp | PGD-BERT | 254 | **0.92** |
|  | BERT-Attack [6] | 273 | 0.77 |
|  | Alzantot [5] | 6137 | -- |
| IMDB | PGD-BERT | 358 | **0.94** |
|  | BERT-Attack [6] | 454 | 0.86 |
|  | Alzantot [5] | 6493 | -- |
| AG | PGD-BERT | 178 | **0.94** |
|  | BERT-Attack [6] | 213 | 0.63 |
|  | Alzantot [5] | 3495 | -- |
| Fake | PGD-BERT | 968 | **0.93** |
|  | BERT-Attack [6] | 1558 | 0.81 |
|  | Alzantot [5] | 28508 | -- |

TABLE III. PERFORMANCE ON THE NATURAL LANGUAGE INFERENCE TASKS- ACCURACY AND PERTURBATION PERCENTAGE

| Dataset | Attack Method | Accuracy (In Absence of Attack) | Accuracy (In Presence of Attack) | Perturb % |
|---|---|---|---|---|
| SNLI | PGD-BERT | 89.4 (H/P) | **3.2/12.6** | **8.2/6.3** |
|  | BERT-Attack [6] |  | 7.4/16.1 | 12.4/9.3 |
|  | Alzantot [5] |  | 14.7/-- | 20.8/-- |
| MNLI Matched | PGD-BERT | 85.1 (H/P) | **5.3/10.8** | **7.4/6.7** |
|  | BERT-Attack [6] |  | 7.9/11.9 | 8.8/7.9 |
|  | Alzantot [5] |  | 21.80/-- | 18.2/-- |
| MNLI Mismatched | PGD-BERT | 82.1 (H/P) | **5.1/10.6** | **7.2/7.0** |
|  | BERT-Attack [6] |  | 7/13.7 | 8.0/7.1 |
|  | Alzantot [5] |  | 20.9/-- | 19.0/-- |

TABLE IV. PERFORMANCE ON THE NATURAL LANGUAGE INFERENCE TASKS – QUERY NUMBER AND SEMANTIC SIMILARITY

| Dataset | Attack Method | Query Number | Semantic Similarity |
|---|---|---|---|
| SNLI | PGD-BERT | **12/18** | **0.75/0.82** |
|  | BERT-Attack [6] | 16/30 | 0.40/0.55 |
|  | Alzantot [5] | 613/-- | -- |
| MNLI Mached | PGD-BERT | **12/21** | **0.72/0.87** |
|  | BERT-Attack [6] | 19/44 | 0.55/0.68 |
|  | Alzantot [5] | 692/-- | -- |
| MNLI Mismached | PGD-BERT | **17/28** | **0.70/79** |
|  | BERT-Attack [6] | 24/43 | 0.53/0.69 |
|  | Alzantot [5] | 737/-- | -- |

TABLE V. PERFORMANCE ON TRANSFERABILITY – ACCURACIES OF OTHER MODELS UNDER BERT-ATTACK AND PGD-BERT

| Dataset | Model | Original Acc | Attacked Acc Under BERT | Attacked Acc Under PGD-BERT |
|---|---|---|---|---|
| Yelp | BERT-Large | 97.9 | 8.2 | 6.1 |
|  | Word-LSTM | 96.0 | 1.1 | 0.8 |
| IMDB | BERT-Large | 98.2 | 12.4 | 9.7 |
|  | Word-LSTM | 89.8 | 10.2 | 8.6 |
| MNLI Mached | BERT-Large | 86.4 | 13.2 | 11.4 |
|  | ESIM | 76.2 | 9.6 | 8.3 |

Tables I - IV present the performance results of our proposed PGD-BERT-Attack models for natural language inference and text classification tasks on different datasets. The results of PGD-BERT are also compared with those of BERT-Attack [6]. and the attack method of Alzantot et al [5]. For NLI tasks, following the approach used in the BERT-Attack [6], the hypotheses (H) and premises (P) are attacked separately. The four metrics used in analyzing the performance of the attack methods are (1) accuracy of classification or inference without and with the attack, (2) perturbation percentage, (3) number of queries to launch the attack, and (4) semantic resemblance between the adversarial and the source texts. It is observed from Tables 1- IV, that our proposed approach, PGD-BERT-Attack has outperformed both the BERT-Attack proposed by Li et al. [6], and the adversarial attack proposed by Alzantot et al. [5], on all metrics and for both natural language inference and text classification tasks. It is also observed that the classification of reviews is an easier task since for all such tasks the percentage of perturbation is very low.

TABLE VI. PERFORMANCE ON TRNSFERABILITY – PERTURBATION PERCTAGE ON OTHER MODELS UNDER BERT-ATTACK AND PGD-BERT

| Dataset | Model | Perturb % Under BERT | Perturb % Under PGD-BERT |
|---|---|---|---|
| Yelp | BERT-Large | 4.1 | 4.0 |
|  | Word-LSTM | 4.7 | 3.2 |
| IMDB | BERT-Large | 2.9 | 1.7 |
|  | Word-LSTM | 2.7 | 2.4 |
| MNLI Mached | BERT-Large | 7.4 | 7.2 |
|  | ESIM | 21.7 | 16.7 |

We have mentioned earlier that the adversarial examples generated using PGD have better transferability meaning they are effective against not only the BERT model but also other models. To compare PGD-BERT-Attack with BERT-Attack on the transferability metric, the attack is launched on some other models, Word-LSTM and BERT-Large, and text classification tasks on IMDB and YELP datasets. Similarly, the MNLI dataset is used for the NLI task. The performance of the BERT-Attack and PGD-BERT-Attack are presented in Table V presents the comparative results of BERT and PGD-BERT on their attack accuracies on two models and three datasets. It is evident that PGD-BERT has more adverse effect on the accuracies for every model-dataset combination. The accuracies of for each model-dataset has a lower accuracy under PGD-BERT in comparison to BERT-Attack.

Table VI presents the relative performance results of BERT-Attack and PGD-BERT based on the perturbation percentages for the same model-dataset combinations. A lower value of perturbation percentage implies a more efficient attack. Again, it is observed that for all model-dataset combination, PGD-BERT's perturbation percentages are lower than the corresponding figures for BERT. Since the percentage of perturbation is smaller for the PGD-BERT for all cases, it is evident that the attack method is efficient as it needs fewer words to change to launch a successful attack.

The results clearly indicate superior ability of PGD-BERT on transferability to other models.

As mentioned in Section IV, our proposed attack scheme PGD-BERT allows for adaptive perturbation by iteratively refining the adversarial perturbation exploiting the feedback from the victim BERT model. If $y_{pred}$ (i.e., the predicted output) is the desired incorrect output, then it continues with the current perturbation. If, on the other hand, $y_{pred}$ is not the desired output, then it adjusts the perturbation to steer the prediction toward the desired output. This adaptive adjustment allows PGD-BERT to craft more effective adversarial examples tailored to the behavior of the BERT language model.

We compared the performance of PGD-BERT's adaptive perturbation ability with the 'flexible candidates choice' feature of BERT-Attack. The word substitution strategy of BERT-Attack has a hyperparameter $K$ that denotes the number of possible words that may substitute a word in the source text to launch the attack. While a large value of $K$ allows for more flexibility, it also leads to less semantic similarity. While it has been observed in the BERT attack that with larger $K$ values semantic similarities fall only by a margin of 2%, the perturbation rates remain considerably higher. For PGD-BERT, on the other hand, with a higher $K$, while the semantic similarity is found to have fallen by a minuscule margin of less than 0.5%, the perturbation rate is also found to be much lower as presented in Table I.

To study the effect of fixed and adaptive $K$ with a pre-set threshold on the attack accuracy and the number of queries made by the attack on a dataset, the performances of PGD-BERT and BERT-Attack are compared on the IMDB and the AG News datasets.

TABLE VII. BERT-ATTACK PERFORMANCE ON THE IMDB DATASET WITH AND FIXED $K$ AND ADAPTIVE K WITH A THRESHOLD VALUE

| Method | Accuracy in Absence of Attack | Accuracy in Presence of Attack | Number of Queries |
|---|---|---|---|
| Fixed K | 90.9 | 11.4 | 454 |
| Variable K with a Threshold | | **12.4** | **440** |

TABLE VIII. PGD-BERT PERFORMANCE ON THE IMDB DATASET WITH FIXED $K$ AND ADAPTIVE K WITH A THRESHOLD VALUE

| Method | Accuracy in Absence of Attack | Accuracy in Presence of Attack | Number of Queries |
|---|---|---|---|
| Fixed K | 90.9 | 8.2 | 358 |
| Variable K with a Threshold | | **9.4** | **340** |

TABLE IX. BERT-ATTACK PERFORMANCE ON THE AG NEWS DATASET WITH AND FIXED $K$ AND ADAPTIVE K WITH A THRESHOLD VALUE

| Method | Accuracy in Absence of Attack | Accuracy in Presence of Attack | Number of Queries |
|---|---|---|---|
| Fixed K | 94.2 | 10.6 | 213 |
| Variable K with a Threshold | | **14.4** | **202** |

Tables VII and VIII present the performance of BERT-Attack and PGD-BERT with a fixed K and a variable $K$ with a threshold on the IMDB dataset. It is observed that for BERT-Attack, with a variable $K$ having a threshold yields a marginally higher accuracy of the classifier but an appreciably fewer number of queries. This indicates that the candidate word list created by the masked language model based on their information scores had some useless words. Not considering those words in the substitution strategy with a variable $K$ having a threshold has yielded fewer queries to launch the attack. The same observations are also found in Table VIII. However, PGD-BERT is more effective in reducing the accuracy of the classification model while requiring fewer queries on the IMDB dataset.

TABLE X. FIXED $K$ AND ADAPTIVE K WITH A THRESHOLD VALUE

| Method | Accuracy in Absence of Attack | Accuracy in Presence of Attack | Number of Queries |
|---|---|---|---|
| Fixed K | 94.2 | 7.4 | 178 |
| Variable K with a Threshold | | **8.7** | **169** |

Tables IX and X present the performance of BERT-Attack and PGD-BERT with a fixed K and a variable $K$ with a threshold on the AG News dataset. On the AG News dataset too, PGD-BERT has outperformed BERT-Attack.

VI. CONCLUSION

This paper introduced a new scheme of text attack on BERT models using projected gradient descent (PGD). Through comprehensive experimentation and analysis, we demonstrated the superior performance of our PGD-based attack in various aspects compared to the standard BERT attack method. Firstly, our PGD-based attack showed a significantly higher attack accuracy, highlighting its effectiveness in generating adversarial examples that successfully mislead BERT models. This increased accuracy underscores the potency of PGD in finding more impactful perturbations in the input space. Secondly, our approach achieved a lower percentage of perturbation than traditional BERT attacks. This indicates that PGD can craft adversarial samples with minimal alterations to their original counterparts, making them less perceptible while still capable of fooling the model. Moreover, our PGD-based attack required fewer queries to the victim model, reducing the computational cost of generating adversarial samples. This efficiency is crucial for real-world applications with limited computational resources or time constraints. Furthermore, our analysis revealed that adversarial examples generated by PGD maintained higher semantic similarity with the original inputs than those generated by traditional methods. This is a crucial advantage as it ensures the perturbed inputs remain contextually relevant, enhancing the stealthiness of the attack. Lastly, our PGD-based attack demonstrated superior transferability across different target models, indicating its robustness and generalizability. This ability to transfer adversarial examples across models highlights the efficacy and practicality of PGD-BERT in real-world applications where diverse models are employed.

In summary, our proposed PGD-based BERT attack presents a significant advancement in adversarial attacks, offering higher accuracy, lower perturbation rates, reduced query requirements, improved semantic similarity, and enhanced transferability. These findings underscore the potential of PGD as a powerful tool for generating adversarial samples in NLP tasks, paving the way for more robust defense mechanisms and improved model robustness against adversarial attacks. Involving human evaluation metrics for evaluation of the performance and adaptive perturbation threshold to enhance the efficacy of PGD-BERT-Attack constitutes some future areas of work.